%% file: 25_chris_usd.tex
\PassOptionsToPackage{pdftex, xetex}{graphicx}
\PassOptionsToPackage{dvipsnames,svgnames,table}{xcolor}
\documentclass[letterpaper, 10pt, conference]{ieeeconf}
\IEEEoverridecommandlockouts
\makeatletter
\let\NAT@parse\undefined
\makeatother
\usepackage[sort,numbers,compress]{natbib}
\makeatother

\input{macros}


\graphicspath{{./figures/}}

\usepackage[color=gray!30,textsize=tiny]{todonotes}

\usepackage{fancyhdr}
\usepackage{algpseudocode}
\usepackage{listings}
\usepackage{ifthen}
\usepackage{censor}

\makeatletter
\DeclareMathSizes{9.95}{9}{7}{5}
\DeclareMathSizes{10}{9}{7}{5}
\DeclareMathSizes{11}{10}{8}{6}
\makeatother

\makeatletter
\DeclareMathSizes{\@xpt}{9}{7}{5}
\makeatother

\title{\bf  Asset-Centric Metric-Semantic Maps of Indoor Environments}

\author{Christopher D. Hsu$^{1,2}$ and Pratik Chaudhari$^{1}$
\thanks{
$^{1}$Department of Electrical \&  Systems Engineering and
General Robotics, Automation, Sensing and Perception (GRASP) Laboratory at the University of Pennsylvania.
\href{mailto:chsu8@upenn.edu}{chsu8@upenn.edu},
\href{mailto:pratikac@upenn.edu}{pratikac@upenn.edu}}
\thanks{
$^{2}$DEVCOM Army Research Laboratory
\href{mailto:christopher.d.hsu.civ@army.mil}{christopher.d.hsu.civ@army.mil}
}
}

\begin{document}

\maketitle

\thispagestyle{fancy}
\fancyhead{} 
\fancyfoot{} 
\fancyhead[C]{\ifthenelse{\value{page}=1}{UNCLASSIFIED \\ Distribution Statement A. Approved for public release: distribution is unlimited.}{UNCLASSIFIED}}
\fancyfoot[C]{UNCLASSIFIED}
\pagestyle{fancy}
\setlength{\headheight}{12pt}
\setlength{\footskip}{12pt}

\begin{abstract}

Large Language Models (LLMs) can help robots reason about abstract task specifications.
This requires augmenting classical representations of the environment used by robots, such as point-clouds and meshes, with natural language-based priors.
There are a number of approaches to do so in the existing literature.
While some navigation frameworks leverage scene-level semantics at the expense of object-level detail, others such as language-guided neural radiance fields (NeRFs) or segment-anything 3D (SAM3D) prioritize object accuracy over global scene context.
This paper argues that we can get the best of both worlds.
We use a Unitree Go2 quadruped with a RealSense stereo camera (RGB-D data) to build an explicit metric-semantic representation of indoor environments.
This is a scene-scale representation with each object (e.g., chairs, couches, doors, of various shapes and sizes) represented by a detailed mesh, its category, and a pose.
We show that this representation is more accurate than foundation-model-based maps such as those built by SAM3D, as well as state-of-the-art scene-level robotics mapping pipelines such as Clio (Maggio et al., 2024).
Our implementation is about 25$\times$ faster than SAM3D and is about 10$\times$ slower than Clio.
We can also adapt our approach to enable open-set scene-level mapping, i.e., when object meshes are not known a priori, by building upon SAM3D to further improve precision and recall.
We show how this representation can be readily used with LLMs such as Google's Gemini to demonstrate scene understanding, complex inferences, and planning. 
We also display the utility of having these representations for semantic navigation in simulated warehouse and hospital settings using Nvidia's Issac Sim. 
\end{abstract}

\input{intro}

\input{related}
\input{method}
\input{experiments}
\input{conclusion}


\begingroup
\footnotesize
\bibliographystyle{IEEEtran}
\bibliography{references, pratik}
\endgroup

\end{document}

%% file: macros.tex
\usepackage{amsthm, amssymb, amsmath, bm, mathtools, physics, dsfont}
\usepackage{pifont}
\usepackage{newtxtext}
\usepackage[dvipsnames,svgnames,table]{xcolor}
\usepackage[pdftex,xetex]{graphicx}
\usepackage{caption}
\usepackage[final]{microtype}

\usepackage[allcolors=Cerulean,colorlinks=true]{hyperref}
\usepackage{url}
\usepackage{footnotebackref}
\usepackage{tablefootnote}
\usepackage{footmisc}
\usepackage{tikz}
\usepackage{bigdelim}
\usepackage{siunitx}
\usepackage{adjustbox}

\usepackage{setspace,float, tabularx, multirow, subcaption, environ, wrapfig, nicefrac,booktabs,colortbl,soul,comment, blindtext}
\usepackage[normalem]{ulem}

\usepackage[capitalize]{cleveref}
\renewcommand{\Cref}[1]{\cref{#1}}
\crefname{section}{Sec.}{Secs.}
\crefname{appendix}{Sec.}{Secs.}

\newcommand{\beq}[1]{\begin{equation}#1\end{equation}}

\theoremstyle{definition}
\newtheorem{theorem}{Theorem}

\newtheorem{remark}[theorem]{Remark}

\def \s {\sigma}

\usepackage[breakable, skins]{tcolorbox}
\usepackage{fontawesome5}
\usepackage{listings}
\usepackage{framed}



%% file: intro.tex

\section{Introduction}
\label{sec:intro}

For tasks ranging from search and rescue, industrial automation, and guiding people through complex environments like airports or museums, a majority of robotic systems use metric representations of the environment for mapping and navigation. Humans use more abstract representations for such tasks. Structures such as place cells and grid cells in the brain \cite{moserPlaceCellsGrid2008,haftingMicrostructureSpatialMap2005} encode space efficiently \cite{weiPrincipleEconomyPredicts2015} and are tuned to sensory and contextual cues \cite{wangTimeMakesSpace2024,wang2025remi}.
These structures have also been argued to represent semantics for more general kinds of decision making \cite{kaplan2019entorhinal}.
The hallmark of human representations of space is indeed their semantic richness---often in lieu of metric details. It stands to reason that robots will benefit by combining semantic information with metric information for building effective representations of space.

This perspective is also a popular line of thought in recent literature. Bootstrapped by foundation models, there are a number of techniques to augment metric representations such as voxel grids, point-clouds, etc. using language-based priors to create metric-semantic maps.
These approaches do work well but they are often tailored to specific tasks, e.g., vision-language models for navigation, language-guided neural radiance fields for mapping, etc. However, these methods lack realistic fine-grained geometry. See \cref{sec:related}.

On the other end of the spectrum, generative models such as SAM3D \cite{sam3dteam2025sam3d3dfyimages} or SceneGen~\cite{meng2025scenegensingleimage3dscene} excel at synthesizing high-fidelity objects within a localized, bounded scene area. While the latter can leverage multiple images for better object quality, they are not designed for large-scale SLAM-like reconstruction, i.e., they cannot integrate long-sequence trajectories to form a continuous, wide-area environment. Conversely, multi-view generative models~\cite{wang2025vggtvisualgeometrygrounded, lin2025depth3recoveringvisual} 
address large-scale scene synthesis by prioritizing global spatial consistency across camera trajectories and multi-view imagery. However, these models typically produce monolithic representations, such as a single continuous mesh or a global radiance field, where objects are geometrically fused with the background. The lack of instance-level segmentation in these representations creates a significant gap for robotics applications, which require the ability to isolate and semantically identify individual objects.

\begin{figure}
    \centering
    \includegraphics[width=\linewidth]{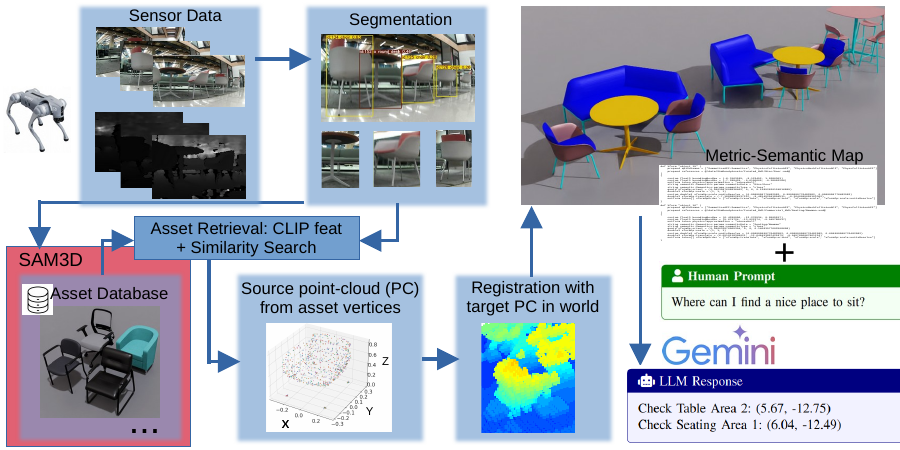}
    \caption{We show our pipeline where a robot collects sensor measurements (RGB-D and odometry) and does semantic segmentation. With each image crop, it retrieves the most similar object from our database, whether generated by SAM3D-objects (red box) or previously collected, simulates the object, and does registration to a global point-cloud. After some reconciliation, we obtain a written metric-semantic map that is fed to an LLM as context to get waypoints such that a robot can navigate to a language-guided goal.}
    \label{fig:real2usd}
\end{figure}


\begin{framed}
We argue that a synergy between classical mapping over trajectories with multi-view observations in conjunction with performant generative models is necessary to create fine-grained scene-level metric-semantic maps. When written as a JSON or an universal scene description (USD) such representations capture scene-level details which can  provide context to LLMs for robot autonomy tasks. To that end, the contributions of this work are (shown in \cref{fig:real2usd}) are as follows.
We demonstrate a system using a Unitree Go2 quadruped robot carrying a RealSense stereo camera that (i) builds an explicit metric-semantic map representation of indoor environments with diverse fine-grained objects, (ii) augments and improves the objects obtained from a generative model called SAM3D to work at the scene-level, and (iii) parses the text map using Google’s Gemini to demonstrate scene understanding, complex inferences, and planning.
\end{framed}

%% file: related.tex
\section{Related Work}
\label{sec:related}

\textbf{3D Metric-Semantic Scene Graph}-based approaches first construct a geometric representation of the scene using simultaneous localization and mapping (SLAM) on 3D voxels constructed from signed distance fields~\cite{oleynikova2017voxblox} or newer techniques like neural radiance fields (NeRFs)~\cite{mildenhall2020nerf} and Gaussian splatting (3DGS)~\cite{kerbl20233dgaussiansplattingrealtime}. This geometric map can be modified to incorporate information about object categories and semantic labels such as book, desk, etc.~\cite{rosinol2021kimeraslamspatialperception, liu2024slideslamsparselightweightdecentralized}.
While some approaches represent this information implicitly, e.g., a instance/category-wise segmentation of a point-cloud, some others abstract the map into a scene graph~\cite{Armeni-2016} with nodes and edges that connect the geometry to higher level concepts such as categories and relationships~\cite{maggio2024cliorealtimetaskdrivenopenset, rosinol2021kimeraslamspatialperception, hughes2022hydrarealtimespatialperception}. Some methods also use LLMs~\cite{gu2023conceptgraphsopenvocabulary3dscene, Rana-SayPlan} to model explicit relationships between objects in the scene.
Approaches such as SLAM++~\cite{Salas-Moreno-Slam++} optimize a pose graph using a limited set of manually scanned objects.
These ``metric-semantic'' representations can support planning and task execution~\cite{ong2025atlasnavigatoractivetaskdriven,Rana-SayPlan} over long horizons~\cite{anwar2024remembrbuildingreasoninglonghorizon}.
 
In our experimental section~\cref{sec:exp}, we will compare our method's metric-semantic mapping accuracy against that of Clio~\cite{maggio2024cliorealtimetaskdrivenopenset} which showed to outperform other metric-semantic works on open-set object mapping~\cite{gu2023conceptgraphsopenvocabulary3dscene, jatavallabhula2023conceptfusionopensetmultimodal3d}. Clio is a real-time framework that uses a stereo camera to build task-driven 3D scene graphs, also see~\cite{Armeni-2016}. It produces 3D bounding boxes around task relevant objects using an information-theoretic objective for agglomerative clustering of CLIP features over unstructured meshes. We will show that at the cost inference time, our resulting map is more accurate than that of Clio. There is an added benefit to our method. As opposed to unstructured meshes like that of Clio or explicit 3DGS/point-cloud representations generated by VGGT~\cite{wang2025vggtvisualgeometrygrounded} or Depth Anything 3~\cite{lin2025depth3recoveringvisual}, our metric-semantic map is composed of explicit and structured fine-grained geometric assets and our process is reminiscent of a real to simulation pipeline.

\textbf{Real to Sim to Real}
was first studied as ``virtualized reality'' by Takeo Kanade and colleagues in 1997 to create immersive experiences of real-world events using multi-baseline stereo across 51 cameras \cite{Kanade-1997-14289}. This idea has again become important because exploiting simulation-based methods in robotics requires that we have effective methods to build realistic virtual versions of physical scenes. For example, \cite{li2024-RoboGSim} use 3DGS to reconstruct the robot workspace as a mesh and import it into a physics simulator. The authors in~\cite{torne2024-rialto} improve upon this to introduce a manual tool to segment the 3DGS representation. Works like \cite{dai2024-ACDC} take a different approach. They import articulated assets into a 3D simulation using a single RGB image as input to obtain very impressive results; others use text prompts \cite{lin2025proc4gemfoundationmodelsphysical} for synthesizing procedural scenes. ShapeR~\cite{siddiqui2026shaperrobustconditional3d} uses Aria glasses for scene reconstruction via volumetric primitives. Works like \cite{yao2025castcomponentaligned3dscene, ni2024phyreconphysicallyplausibleneural} use simulators to aid the generation of assets.

Our method builds on SAM3D-objects~\cite{sam3dteam2025sam3d3dfyimages}, which reconstructs meshes or 3DGS from single RGB(-D) image and associated object masks. While SAM3D effectively ``hallucinates'' unobserved geometry and texture via generative priors, its reliance on learned flow matching~\cite{lipman2023flowmatchinggenerativemodeling} for pose prediction—rather than explicit optimization—can introduce prediction errors. We extend SAM3D’s capabilities from single images to continuous streams, correcting pose inaccuracies and enabling full-scene reconstruction. Our mapping solution is a fast addition (in spite of multiple large models being queried at each instant), remains view-consistent despite occlusions, and integrates metric and semantic data to provide essential scene context for LLM-powered robotic tasks.


%% file: method.tex

\section{Methods}
\label{sec:method}

\cref{fig:real2usd} shows our entire pipeline. We develop a method to construct metric-semantic representations from sensory data collected by robots in real world indoor environments.
There are three main components: recognition and retrieval which involves using image segmentation models to identify and robustly query a database of objects or build detailed meshes of new objects on the fly (\cref{ssec:recognition_and_retrieval}), placing these objects into the scene at their correct pose and sizes (\cref{ssec:localization}) and reconciliation to ensure that the entire scene is physically plausible (\cref{ssec:reconciliation}). Alongside, we use a JSON or a universal scene description (USD) to reference individual object representations and their pose information in the scene.%
\footnote{
The narrative will contain footnotes that describe some problem-dependent ``tricks'' that were used to ensure a performant and reproducible pipeline.
}

We assume that the robot has an accurate estimate of its pose $s_t \in \text{SE}(3)$ at time $t$, i.e., there exists a system for localization, or odometry, with known transformations to all sensors. Suppose the robot obtains observations $y_1^t = (y_1, \dots, y_t)$ from viewpoints $s_1^t$ respectively. An observation consists of $y_t \equiv (y_t^{\text{rgb}}, y_t^{\text{depth}})$ which are the RGB image and depth data obtained from a stereo camera system (Intel RealSense), from a viewpoint $s_t$. The depth image is transformed into a point-cloud and registered against a global frame of reference. We want to build a representation of a static scene, denoted by $\xi_t$, using these observations. Suppose we have an initial database of objects $D = \{\xi^{(1)}, \dots, \xi^{(n)}\}$ (chairs, tables, doors, etc.) represented as meshes (GLBs), textured computer-aided-design (CAD) models, or universal scene description assets (USDs). We will also discuss how during the course of the experiment, $D$ can be augmented to include new objects using generative models like SAM3D-objects~\cite{sam3dteam2025sam3d3dfyimages} such that $D = \{\xi^{(1)}, \dots, \xi^{(n)}, \dots, \xi^{(m)}\}$.
We will represent the scene using these objects, i.e.,
\beq{
    \xi_t \equiv \{ (\xi^{(i)}, x^{(k)} ) \}_{i \in \{1,\dots, m\}, k\in \{1,\dots, k_t\}}.
    \label{eq:scene}
}
There are $k_t \equiv \abs{\xi_t}$ total instances of objects at poses $x^{(k)} \in \text{SE}(3)$ in the global frame corresponding to track ID $k$. This $k$ denotes a spatial ``slot'' $S_k$ that we seek to fill with the best object $\xi^{(i)}$. There can be multiple instances of the same object $\xi^{(i)}$ in $\xi_t$ but at different slots at poses $x^{(k)}$. 



\subsection{Object Recognition and Retrieval}
\label{ssec:recognition_and_retrieval}

The representation of the scene $\xi_t$ should explain the observations in $y_1^t$, i.e., a rendering from the scene representation should have the correct photometric and geometric properties.%
\footnote{We are interested in semantic navigation tasks in this paper, and therefore it is not important to capture the photometric properties extremely precisely.}
We now describe how to recognize and retrieve semantically-similar objects for constructing the scene.

\begin{figure} [!htbp]
    \centering
    \includegraphics[width=0.4\linewidth]{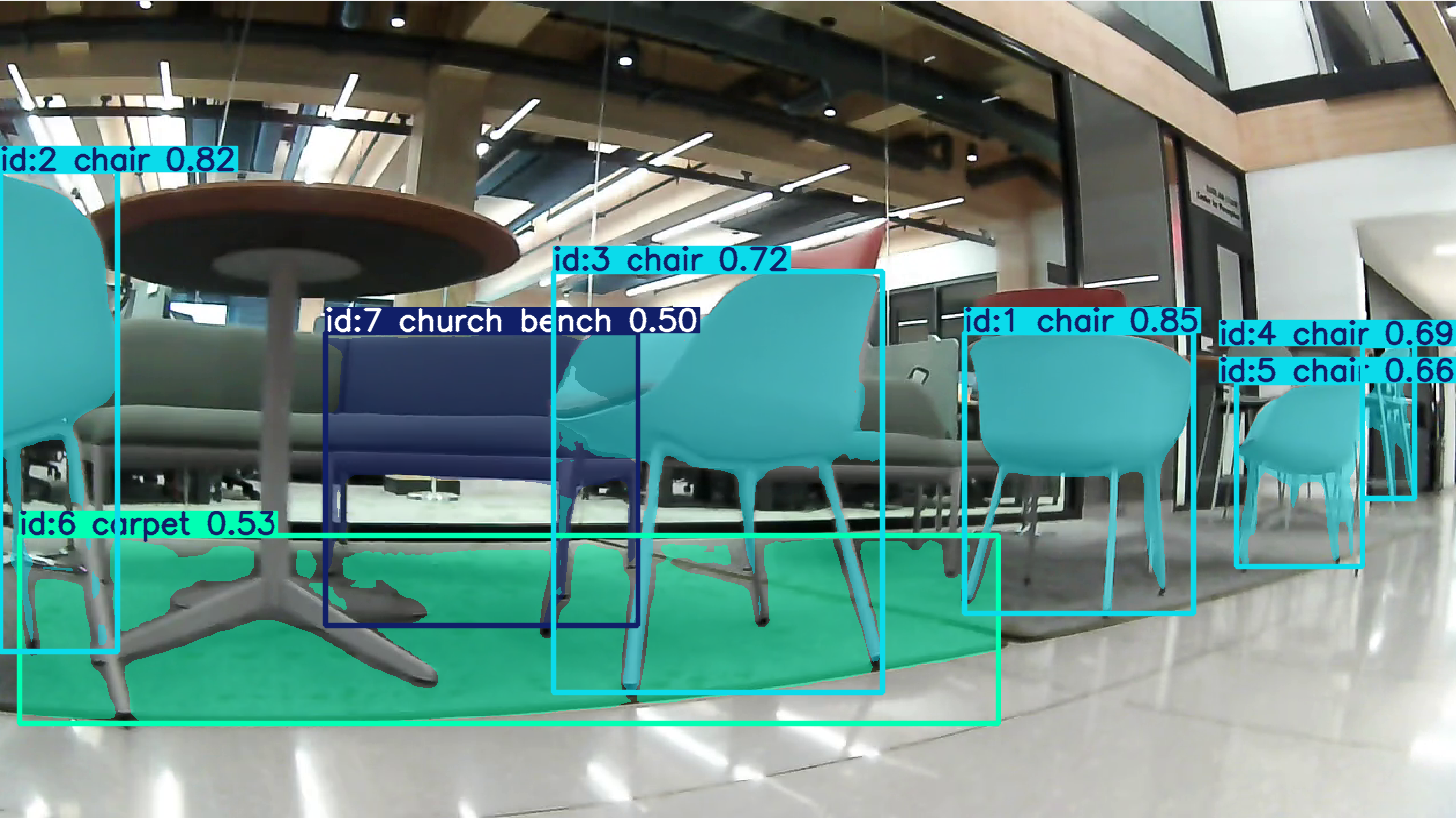}
    \includegraphics[width=0.4\linewidth]{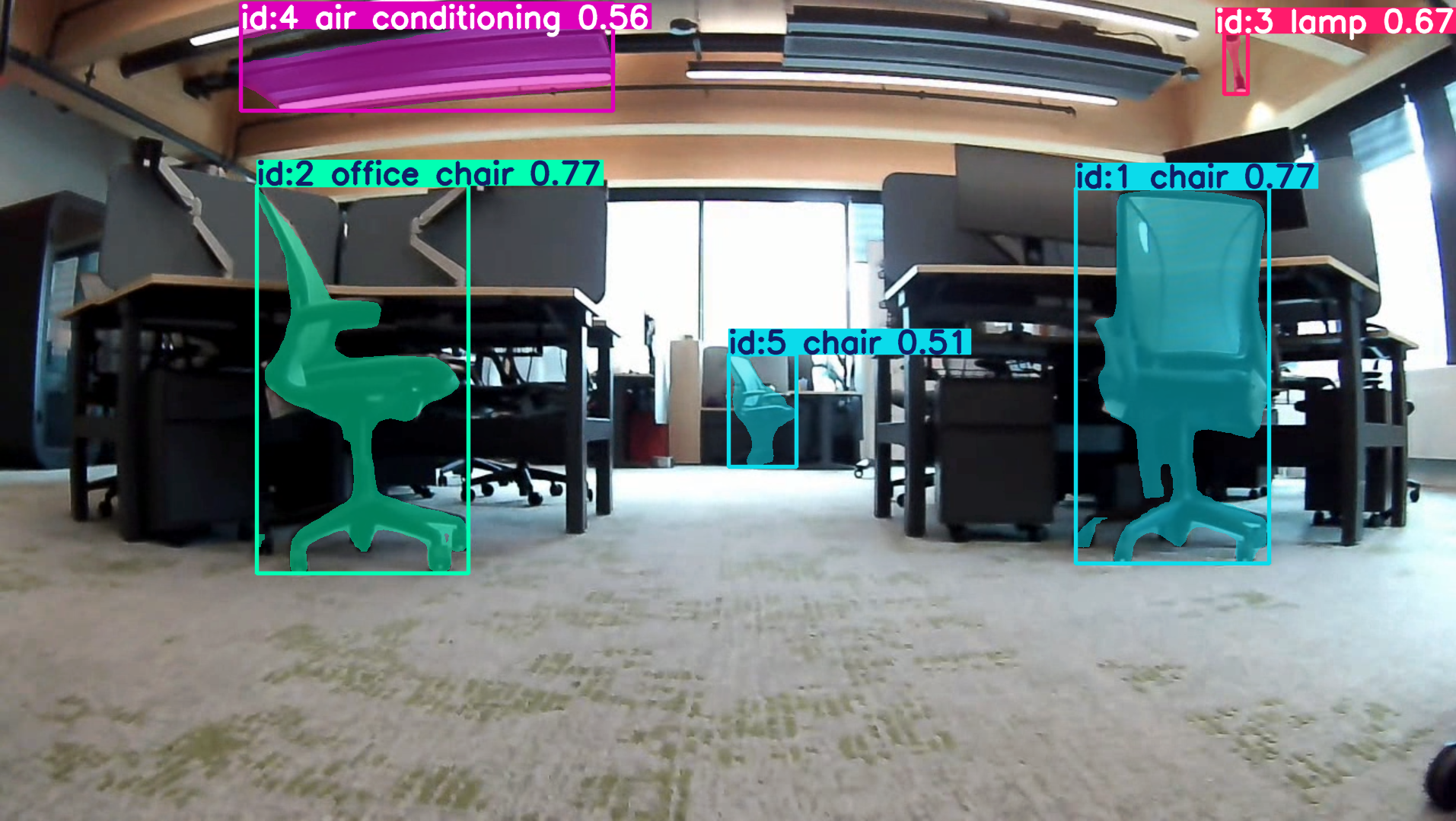}
    \caption{
    \textbf{An example of inadequate precision and recall in open-set detection models such as YOLOE.}
    In the left image, the couch is typically categorized as a ``church bench'' (dark blue).
    We use heuristics to map such fine-grained but semantically incorrect labels to more general labels, e.g., the ``church bench'' is modified to be a ``chair''.
    Notice that the round table and desks are not segmented in these images, even with prompts such as ``table'' or ``desk''. This inadequacy is problematic for robotics pipelines that have much more stringent requirements on their performance. We work around this by querying the model twice (see the text).
    }
    \label{fig:segmentation}
\end{figure}

\textbf{Robust object detection with open-set models.}
Given an observation, $y_t^{\text{rgb}}$ we compute a set of putative 2D object masks, their labels using YOLOE~\cite{wang2025yoloerealtimeseeing} shown in~\cref{fig:segmentation}, and their tracking ID (which can also be obtained from YOLOE)~\cite{aharon2022botsortrobustassociationsmultipedestrian}.
This model can perform open-set object detection and segmentation by taking a text prompt as input.
Textual prompts are useful to guide the detections onto a specific set of objects, but in indoor environments with many different objects, it is not easy to define text prompts with good recall.
We therefore first run YOLOE without text prompts to use its open-set functionality.
This provides us a set of putative labels which are then used to run the prompt-based YOLOE to obtain detections. This strategy results in twice as many calls to the detection model, but we found that doing so can ensure a high recall in challenging environments.
In contrast to using Segment Anything (SAM)~\cite{kirillov2023segment} and CLIP to retrieve labels~\cite{radford2021learningtransferablevisualmodels}, YOLOE provides semantically consistent masks without over-segmenting objects.


\textbf{Semantic search for retrieving objects from an open-set database.}
Suppose we have masks $m_k$ from the segmentation model, labels $\ell_k$, and tracking ids $k$. For each $k$ we instantiate a spatial slot $S_k\equiv \{s_{k}, m_{k}, \ell_k, P_k \}$ where $P_k$ is the unprojected accumulated point-cloud from the depth observations $y_t^{\text{depth}}$ that are associated with $k$ and a robot viewpoint $s_k$. For each slot slot $S_k$ we seek to define a global pose of the object $x^{(k)}$ and a reference to semantically-labeled object $\xi^{(i)}$. In this section we describe how to retrieve objects $\xi^{(\cdot)}$ from the database $D$ that are semantically and photometrically consistent with these detections. In a slower, separate process, we augment $D$ by generating objects from slots $S_k$ using a generative model called SAM3D that can perform visually grounded 3D object reconstruction from a single image~\cite{sam3dteam2025sam3d3dfyimages}.

First, it is reasonable to ask why we should maintain a database of known objects in view of models like SAM3D which can reconstruct new objects on the fly. The first reason is simply that generative are much too slow for real-time operations in robotics, e.g., it takes about 20-30 seconds (on an Nvidia 5090) for SAM3D to generate a mesh of a single object.
\includegraphics[width=0.24\linewidth]{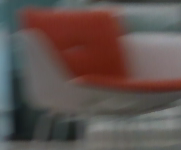}
\includegraphics[width=0.24\linewidth]{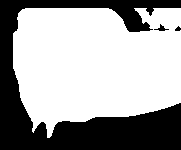}
\includegraphics[width=0.48\linewidth]{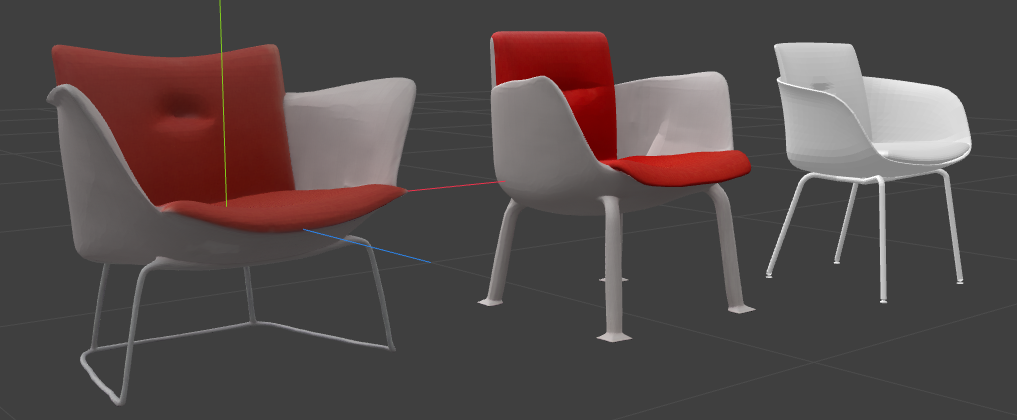}
Even if this were not an issue, the adjoining figure gives an example of an image of a chair from our data (left) and the mask $m_{k}$ (middle). The left-most object instance in the third image (right) shows that the object mesh generated by SAM3D is clearly incorrect in its shape, perhaps due to the viewpoint and visual blur. On the other hand, the retrieval-based procedure that we will describe next shows that we can (i) retrieve a more accurate mesh constructed via objects in the database that were reconstructed at previous times, and (ii) that this is much more faithful to the true object mesh (sourced from the furniture manufacturer). In other words, maintaining a database of reconstructed objects and querying it allows us to fill a slot $S_k$ more robustly and faithfully than single-shot models such as SAM3D.

Therefore, whether we want fast inference (via a previously collected database) or slower, on-demand generated objects, we need to find the closest object that matches our detections.
A naive solution would use the text class label $\ell_{t,k}$ to retrieve the object using, say, CLIP. However, within the general label of a ``chair'' or a ``table'' there are all sorts of different chairs and tables.
We therefore implement an image-based object retrieval procedure that retrieves objects much more robustly.
We use CLIP~\cite{radford2021learningtransferablevisualmodels} to compute embeddings and a similarity search library called FAISS~\cite{johnson2019billion}. In an separate (headless) process, for each object (USD or GLB) in the database $D$, we use Issac Sim (for USD) or PyVista (for GLB) to render 8 views and compute their CLIP embeddings to create a FAISS search index.%
\footnote{We chose views that tend to match the angle for which the robot view the object, i.e., from a lower vantage point.}
%
Object retrieval is performed by embedding the image from a track ID $k$ and searching for the nearest image (in terms of cosine distance) in the search index, and its associated object.
We build in robustness by filtering out results that do not match the queried label. An object track can potentially retrieve multiple labels based on different views, and therefore multiple queries are performed. The reconciliation module in \cref{ssec:reconciliation} will resolve the ambiguity.
\footnote{The authors in \cite{dai2024-ACDC} also use multiple views and CLIP for object retrieval but they also incorporate a vision-language model (VLM). In practice we found this extra step not necessary, and it was in fact a computational bottleneck in the pipeline.}

We built a database of objects using assets in Nvidia's Omniverse SimReady \cite{nvidiaomniverse}. These objects are physically accurate, with realistic physical properties, materials, and meshes. They can be readily used in physical simulations. We also sourced CAD and REVIT models of the furniture in our scenes and converted them into ``Sim Ready'' USD files. We have 27 different types of chairs which range from couches, desk chairs, barstools, and chairs one would find in an office building. We have 24 different types of tables that range from high top tables, office desks, round and square tables, and smaller side tables. We also include 7 miscellaneous objects that would be found in an office such as doors, phone booths (zoom video call pods), power outlets, and monitors.

\subsection{Object Localization}
\label{ssec:localization}


In this step, we focus on placing an object $\xi^{(i)}$ into the scene in a way that is geometrically consistent with the accumulated point-cloud $P_k = \bigcup_{s=1}^t y_{s,k}^{\text{depth}}$ in the sensor data associated with a slot $S_k$. This involves finding an appropriate pose $x^{(k)} \in \text{SE}(3)$ for the object. This is in fact a registration problem and it can be solved using vertices of the mesh/CAD of the object $\xi^{(i)}$.



\textbf{Registration object detections to point-clouds.}
Iterative closest point (ICP) based registration between the object's vertices $\hat y^{(i)}$ and the data $P_k$ results in large amounts of aliasing because the latter can represent a large number of other objects.
However, we can take advantage of the objects segmentations to improve the accuracy of registration.
Assuming known camera intrinsics and the associated robot viewpoint $s_k$ corresponding to the slot $S_k$, we segment the entire slot's point-cloud $P_k$ to select only the depth points that (when projected on the corresponding image plane) lie within object masks $\{m^{(i)}_s \}_{s \leq t}$ where the object was detected. Given this sub-sampled point-cloud and the object $\hat y^{(i)}$, we first use fast global registration~\cite{Zhou_2016_fastglobalregistration} for initialization and then point-to-point ICP~\cite{Besl-p2pregistration} as the second step. The former registration technique does not require initialization but we found that it is not very precise, while the latter technique can perform fine-grained registration with appropriate initialization.%
\footnote{Although we do not do so in this paper because our objects are usually of correct scale, by including a search for scale during ICP, this approach for registration can also handle situations when objects retrieved from the database have a different size than the ones in the point-cloud.}

SAM3D is trained by the original authors to predict the pose of the object in the local camera frame via flow matching~\cite{lipman2023flowmatchinggenerativemodeling} using a monocular image.
This is not very accurate or robust in our real-world experimentation.
We found that it is beneficial to provide depth to this model so that the predicted pose and scale are accurate. Furthermore, we will see in~\cref{sec:exp} that the registration process improves the localization of the pose $x^{(k)}$ of an object $\xi^{(i)}$.


\subsection{Object Reconciliation}
\label{ssec:reconciliation}

The previous two sections described a procedure to select objects from a database to explain the sensor observations collected from a robot.
Different elements of this pipeline for detection, retrieval, registration, etc. need to be exceedingly accurate for the representation to be faithful at the scene-level.
Small inaccuracies in pose can create artifacts such as chairs that ``intersect'' with tables, doors that sink into the floor, etc.
The accuracy of existing state-of-the-art detection, segmentation or object reconstruction models is not sufficient to ensure physically plausible predictions \cite{jena2022beyond}.
Our approach represents the scene explicitly in terms of the objects in the database, and we can therefore take advantage of simulation to improve the physical plausibility of the scene.



\textbf{Pruning multiple detections and incorrect registrations.}
Periodically, across time-steps $t$, we construct a $k$-d tree of the registered point-clouds of all $n_t$ objects in the scene $\xi_t$, let us call it $\bigcup_i y^{(i)}_{i \in \{1, \dots, n_t\}}$. This helps facilitate the operations described next.
The registered point-cloud can often contain multiple objects for a single real-world object, e.g., multiple chairs or tables at slightly different poses.
It can also contain situations when a large object from the database, e.g., a couch, is registered to multiple smaller objects in the real-world such as a chair and a table.
We first prune multiple detections using non-maximum suppression (NMS) based on Euclidean position to get clusters of objects. For each cluster, we need to associate a single object that accurately captures the point-cloud and semantics. To do so, we use a ``distribution score'' and a ``density score'' computed using the object, denoted by $\hat y$ here, and its intersection with the point-cloud of the scene $\bigcup_{i \in \{1, \dots, n_t\}} \{\hat y \cap y^{(i)} \}$. These scores are heuristics that are designed to compute the degree to which a particular object can explain the point-cloud corresponding to the scene. Assets within the same cluster without maximal scores are dropped, and thereby each cluster is only represented by a single object with the maximal score.

\begin{figure}[!htpb]
    \centering
    \includegraphics[width=0.25\linewidth]{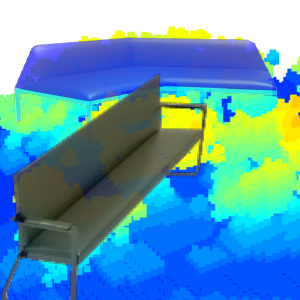}
    \hspace*{1em}
    \includegraphics[width=0.25\linewidth]{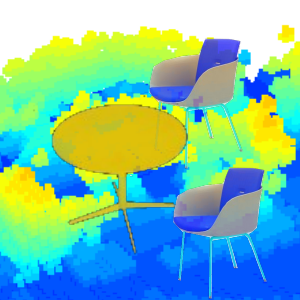}
    \caption{\textbf{Distribution and density scoring.}
    (Left) The blue couch would have a high distribution score due to the point-cloud filling the object uniformly in comparison to the black couch which spans two clusters. (Right) The table and the top chair would have high density scores due to being predominantly filled with points compared to the bottom chair in free space.}
    \label{fig:scoring}
\end{figure}

We show an example of the scoring system in~\cref{fig:scoring} which works as follows. We compute the convex hull of the point-cloud $\hat y$ and discretize it into voxels of size 0.01 \unit{m}.
Let $C$ be the set of voxels with non-zero point-counts object.
The distribution score is the coefficient of variation of these counts which measures the uniformity of the points in the hull. This is computed as follows.
If $\mu, \s$ are the mean and standard deviation of the set $C$, then
the distribution score is $1 - \frac{\sigma}{\mu}$ if $\abs{C} > 0$.
The density score is simply $\abs{C}/\text{total voxels}$.
The distribution score helps disambiguate the couch from the two chairs in the example described above.
The density score helps select objects that have been registered better, i.e., it helps select for better position and orientation matches.

\begin{wrapfigure}{r}{0.3\linewidth}
    \centering
    \includegraphics[width=\linewidth]{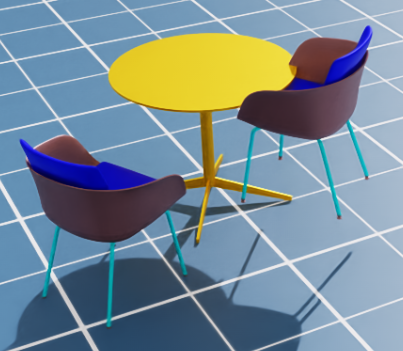} \\
    \vspace*{0.5em}
    \includegraphics[width=\linewidth]{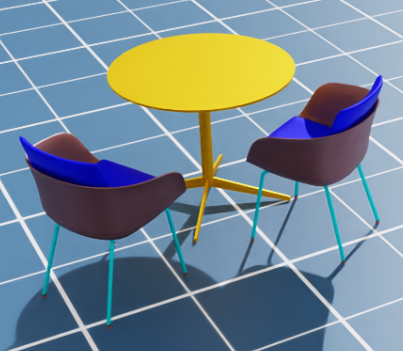}
    \caption{
    (Top) The chair is registered partially into the table. (Bottom) By forward simulating the scene, the chair and table separate and the chair rests on the ground plane.}
    \label{fig:physics}
\end{wrapfigure}
\textbf{Using simulation to obtain a physically plausible static scene.}
Our approach for reconciliation of the retrieved objects helps significantly in constructing a geometrically faithful representation of the scene. But this is not necessarily physically plausible. For example, an object on top of another in the real-world might be represented using an object that floats in mid-air.
We use simulation to improve the physical plausibility, the adjoining figure (\cref{fig:physics}) gives an example.
For all objects, we add rigid body and collision properties in a simulator (Isaac Sim) to instantiate the scene $\xi_t$.
We can then forward simulate the scene to let different objects ``settle'' to their physically plausible locations, e.g., the object floating in mid-air falls down, objects that are in collision with others separate out.%
\footnote{Note that if the forward simulation of the scene causes an object to move to free space, the reconciliation metrics will re-rank the objects in the cluster such that a potentially new object will be the best representative.}
In our experiments, this procedure works well for indoor scenes without moving objects.


%% file: experiments.tex

\section{Experimental Evaluation}
\label{sec:exp}

\subsection{Real World Experiments}

We evaluate our approach against Clio~\cite{maggio2024cliorealtimetaskdrivenopenset} and SAM3D~\cite{sam3dteam2025sam3d3dfyimages} using real world robotic experiments. For quantitative evaluation of our resulting metric-semantic maps, we compare the bounding boxes of objects to those obtained from the baselines based on task descriptions in~\cref{tab:per-class-metric}. As discussed in~\cref{sec:related}, Clio is a framework for building task-driven open-set 3D scene graphs. It produces 3D bounding boxes around task relevant objects by clustering CLIP features over reconstructed meshes. 
Our other baseline, SAM3D, is not a multi-view scene-level method. But we can modify it as follows. We use object detection masks from YOLOE and the robot pose to transform the meshes (GLBs) returned by SAM3D, which are in the camera frame, into the global frame. The label from object detection is attached to each mesh returned by SAM3D, which are composed over time to create a scene-level metric-semantic map. In order to compare all methods, we created the ground truth by manually labeling 3 scenes: small office, a hallway, and a lounge.%
\footnote{We collect an accumulated point-cloud of the scenes and with a 3D bounding box tool, we bound and label the point-clouds with generic labels such as chair or table.}

\begin{figure}[!htpb]
    \centering
    \includegraphics[width=0.85\linewidth]{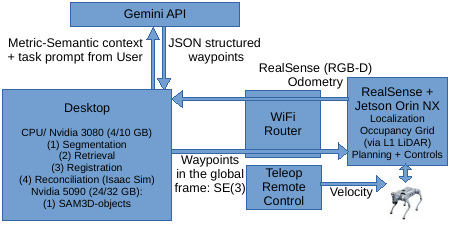}
    \caption{System diagram of our full pipeline.}
    \label{fig:system-diagram}
\end{figure}

\paragraph{Real World Data Collection}
We collect data from three real world indoor scenes: a small office, a hallway, and a lounge, by manually teleoperating a Unitree Go2 quadruped robot. The robot collects RGB images and depth data (via a stereo camera) from a RealSense D435i, and odometry via Go2's default SDK. In our implementation, we stream all data over WiFi via the WebRTC protocol to a desktop (with a Nvidia 3080 GPU running headless Isaac Sim and a 5090 GPU that is used to run SAM3D). A system diagram is show in \cref{fig:system-diagram}.
At the end of data collection, we extract bounding boxes, object meshes, the map etc. to obtain quantitative metrics using manually labeled ground truth objects.%

\paragraph{Quantitative Results of Metric-Semantic Mapping}
We would like to evaluate the effectiveness of the semantic mapping in terms of the accuracy of object locations and their spatial geometry as well as the correctness of the class.
Our manual object categories are different from those of Clio where they depend upon the task. We therefore run Clio with tasks such as ``an image of a \{class\}''. Boxes that are grouped with certain tasks will directly correspond to the ground truth (GT) bounding box labels. As for our method, we label objects by the output of the YOLOE segmentation model. Every object label in our final map is associated to a generic semantic label to enable a fair comparison, e.g., a ``Steelbook desk chair'' and a ``barstool'' are both associated to the label ``chair''.

We focus the following evaluation on three classes: chairs, tables, and doors.%
\footnote{We define a suite of metrics to contextualize quantitative results in our experiments as described next. This is because, metrics such as intersection-over-union (IOU) or accuracy are uniformly poor for existing open-set methods in the literature in realistic indoor environments.}
A prediction is ``relevant'' if its canonical label is the same as the ground truth label. A prediction and ground truth pair is valid if the 3D intersection-over-union $\text{3D IOU} \geq \tau$ where $\tau = 0.01$. We define two accuracy conditions based on centroid placement: strict accuracy requires the centroids of both the prediction and the GT to be within each other's oriented bounding box (OBB), while relaxed accuracy requires only one centroid to be within the other’s OBB. Open-set recall (osR) is calculated by selecting the top $n$ predictions (where $n$ is the number of GT objects for a class) sorted by IOU; the osR is the fraction of these GT objects successfully matched. Similarly, open-set precision (osP) is the fraction of the $n$ predictions that meet both the $\tau$ threshold and the centroid condition. We report strict and relaxed variants for both osR and osP, along with their harmonic mean (F1 score). Finally, mIOU represents the mean 3D IOU across all classes, and ``Objs'' indicates the fraction of predicted objects compared to GT.

``Our'' method denotes the approach in  \cref{sec:method} with a fixed pre-defined object database. ``SAM3D+Ours'' denotes our entire approach but with the additional step where the object database grows incrementally using meshes from SAM3D, as discussed in \cref{sec:method}.

\begin{figure}[!htbp]
    \centering
    \includegraphics[width=0.82\linewidth]{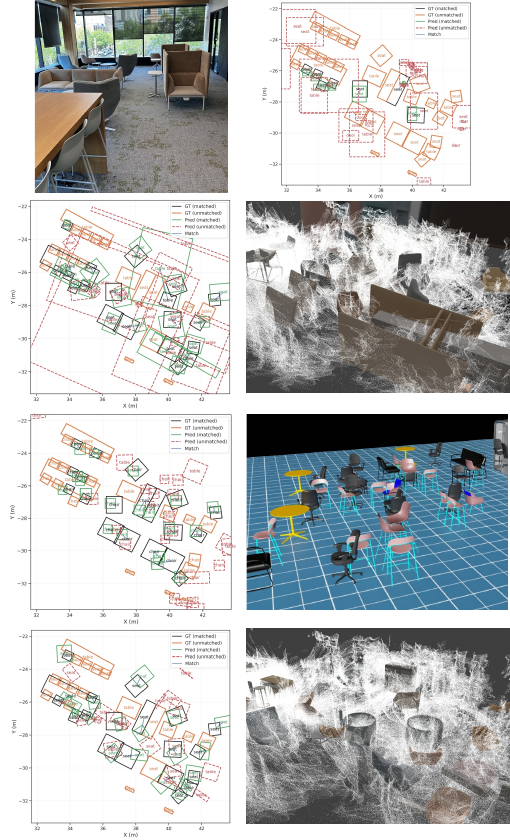}
    \caption{(Top left) We show a real image of the lounge scene viewed from (32, -26). For evaluation, we plot the 2D bounding boxes of the objects of interest (chairs, tables, and doors) of the following experiments: (top right) Clio, (second row) SAM3D, (third row) Ours, and (bottom row) SAM3D+Ours. The ground truth in yellow changes to black if matched (blue line) by label and maximum mIOU to a predicted bounding box (green). Unmatched predictions are denoted as red dotted bounding boxes. Next to the SAM3D-based experiments we show a rendering of the assets against the registered point-cloud. For the SAM3D method (second row right) we have pruned some of the larger badly hallucinated generated objects.}
    \label{fig:qualitative_results}
\end{figure}

In~\cref{tab:per-class-metric} and \cref{tab:overall-metric}, we observe that both Clio and SAM3D have higher accuracies but lower mIOU due to their tendency to produce large bounding boxes, also see~\cref{fig:qualitative_results}. Clio tends to over cluster objects due to the ambiguity between multiple instances of a single class.%
\footnote{For example, a group of chairs is usually clustered into a single large ``chair'' by Clio. This happens because repetitive objects will have near similar CLIP embeddings which when in close proximity, the clustering algorithm will associate as one object.}
Recall of Clio is also rather small. SAM3D, on the other hand, creates extremely large objects when the image only has a partial view. In comparison, we see that our method is able to produce more focused object detections in situations where objects of similar semantic categories are close together. When we combine our method with SAM3D, the process of retrieval prunes away bad generations, registration refines the localization of the object for better mIOU, and reconciliation further fixes bad predictions.

\begin{table}[!htpb]
    \centering
    \LARGE
    \resizebox{1.0\linewidth}{!}{
    \renewcommand{\arraystretch}{1.5}
    \begin{tabular}{l  c | rrrr | rrrr | rrrr}
        \toprule
         &  & \multicolumn{4}{c}{\textbf{chair}} & \multicolumn{4}{c}{\textbf{table}} & \multicolumn{4}{c}{\textbf{door}} \\
        \textbf{Scene} & \textbf{Method} & \textbf{mIOU} $\uparrow$ & \textbf{sAcc} $\uparrow$ & \textbf{rAcc} $\uparrow$ & \textbf{\#obj} & \textbf{mIOU} $\uparrow$ & \textbf{sAcc} $\uparrow$ & \textbf{rAcc} $\uparrow$ & \textbf{\#obj} & \textbf{mIOU}$^*$ $\uparrow$ & \textbf{sAcc}$^*$ $\uparrow$ & \textbf{rAcc}$^*$ $\uparrow$ & \textbf{Objs}\\
        \midrule
        \multirow{4}{*}{small office} & Clio & 0.122 & 0.250 & 0.250 & 2/4 & 0.0 & 0.0 & 0.0 & 0/2 & 0.0 & 0.0 & 0.0 & 1/1 \\
        & SAM3D & 0.204 & \textbf{0.750} & \textbf{0.750} & 3/4 & \textbf{0.001} & 0.0 & 0.0 & 1/2 & \textbf{0.341} & 0.0 & \textbf{1.0} & 1/1\\
        & Ours & \textbf{0.320} & \underline{0.500} & \underline{0.500} & 6/4 & 0.0 & 0.0 & 0.0 & 0/2 & 0.132 & \textbf{1.0} & \textbf{1.0} & 1/1\\
        & SAM3D+Ours & \underline{0.300} & \textbf{0.750} & \textbf{0.750} & 3/4 & 0.0 & 0.0 & 0.0 & 1/2 & \underline{0.258} & 0.0 & \textbf{1.0} & 1/1\\
        \midrule
        \multirow{4}{*}{hallway} & Clio & 0.119 & \textbf{0.273} & \underline{0.606} & 19/33 & \underline{0.109} & 0.167 & \textbf{1.0} & 16/6 & 0.002 & 0.0 & 0.0 & 11/14 \\
        & SAM3D & 0.112 & \textbf{0.273} & \textbf{0.727} & 20/33 & 0.094 & 0.167 & \underline{0.833} & 3/6 & \underline{0.199} & \underline{0.071} & \textbf{0.286} & 6/14\\
        & Ours & \textbf{0.187} & \underline{0.242} & 0.394 & 24/33 & 0.051 & 0.167 & 0.167 & 3/6 & 0.053 & \underline{0.071} & \textbf{0.286} & 4/14\\
        & SAM3D+Ours & \underline{0.149} & \underline{0.242} & 0.303 & 20/33 & \textbf{0.114} & 0.167 & 0.167 & 3/6 & \textbf{0.209} & \textbf{0.143} & \underline{0.143} & 6/14\\
        \midrule
        \multirow{4}{*}{lounge} & Clio & 0.108 & 0.207 & 0.690 & 19/29 & 0.035 & 0.0 & 0.0 & 8/7 & \textbf{0.018} & 0.0 & 0.0 & 9/2\\
        & SAM3D & \underline{0.187} & \textbf{0.586} & \textbf{0.966} & 28/29 & \underline{0.058} & \textbf{0.286} & \textbf{1.0} & 10/7 & 0.0 & 0.0 & 0.0 & 1/2\\
        & Ours & 0.153 & 0.207 & 0.379 & 35/29 & 0.00 & 0.0 & 0.0 & 5/7 & \underline{0.014} & 0.0 & 0.0 & 3/2\\
        & SAM3D+Ours & \textbf{0.197} & \underline{0.483} & \underline{0.759} & 21/29 & \textbf{0.165} & \textbf{0.286} & \underline{0.286} & 4/7 & 0.0 & 0.0 & 0.0 & 1/2\\
        \bottomrule
    \end{tabular}
    }
    \caption{Per class metrics for 3 scenes comparing Clio, SAM3D, Our method, and SAM3D+Our method. We report the mean over the class for mean intersection over union (mIOU), strict accuracy (sAcc), and relaxed accuracy (rAcc). Best and second-best results are bold and underlined, respectively. $^*$ The high aspect-ratio of doors leads to unduly poor IOUs, so we pad the footprint by 0.5 m to ensure that our metrics remain meaningful. }
    \label{tab:per-class-metric}
\end{table}

\begin{table}[!htpb]
    \centering
    \LARGE
    \resizebox{1.0\linewidth}{!}{
    \renewcommand{\arraystretch}{1.5}
    \begin{tabular}{l  c | r r r | r r r | r}
        \toprule
         &  & \multicolumn{3}{c}{\textbf{Strict}} & \multicolumn{3}{c}{\textbf{Relaxed}} &\\
        \textbf{Scene} & \textbf{Method} & \textbf{osR} $\uparrow$ & \textbf{osP} $\uparrow$ & \textbf{F1} $\uparrow$ & \textbf{osR} $\uparrow$ & \textbf{osP} $\uparrow$ & \textbf{F1} $\uparrow$ & \textbf{mIOU} $\uparrow$\\
        \midrule
        \multirow{4}{*}{small office} & Clio & 0.143 & 0.333 & 0.200 & 0.143 & \underline{0.667} & 0.235 & 0.164 \\
        & SAM3D & 0.0 & 0.333 & 0.0 & \textbf{0.571} & \textbf{0.833} & \textbf{0.678} & 0.117 \\
        & Ours &  \textbf{0.429} & \underline{0.429} & \textbf{0.429} & \underline{0.429} & 0.571 & \underline{0.490} & \textbf{0.235} \\
        & SAM3D+Ours & \underline{0.286} & \textbf{0.667} & \underline{0.400} & \textbf{0.571} & \textbf{0.833} & \textbf{0.678} & \underline{0.187}\\
        \midrule
        \multirow{4}{*}{hallway} & Clio & 0.094 & 0.109 & 0.101 & 0.151 & 0.304 & 0.202 & 0.088 \\
        & SAM3D & 0.075 & \underline{0.310} & \underline{0.337} & \underline{0.226} & \underline{0.655} & \underline{0.337} & 0.094 \\
        & Ours & \textbf{0.283} & \textbf{0.600} & \textbf{0.385} & \textbf{0.340} & \textbf{0.767} & \textbf{0.471} & \textbf{0.265} \\
        & SAM3D+Ours & \underline{0.113} & 0.276 & 0.161 & 0.151 & 0.448 & 0.226 & \underline{0.113}\\
        \midrule
        \multirow{4}{*}{lounge} & Clio & 0.053 & 0.111 & 0.071 & 0.105 & 0.194 & 0.137 & 0.136 \\
        & SAM3D & \underline{0.132} & \underline{0.346} & \underline{0.191} & 0.211 & 0.500 & 0.296 & 0.111 \\
        & Ours & \textbf{0.368} & \textbf{0.349} & \textbf{0.358} & \textbf{0.474} & \textbf{0.674} & \textbf{0.557} & \underline{0.148}\\
        & SAM3D+Ours & \underline{0.132} & 0.308 & 0.184 & \underline{0.237} & \underline{0.577} & \underline{0.336} & \textbf{0.154}\\
        \bottomrule
    \end{tabular}
    }
    \caption{Consolidated results for locating objects of interest (chairs, tables, doors) in three environments. The best and second-best results are bold and underlined, respectively. $^*$ The high aspect-ratio of doors leads to unduly poor IOUs, so we pad the footprint by 0.5 m to ensure that our metrics remain meaningful.}
    \label{tab:overall-metric}
\end{table}



\begin{figure}
\centering
\includegraphics[width=0.22\linewidth]{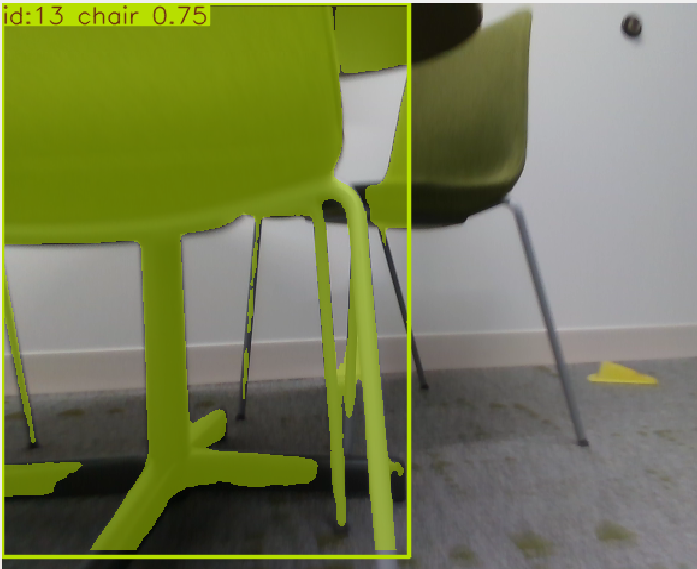}
\includegraphics[width=0.24\linewidth]{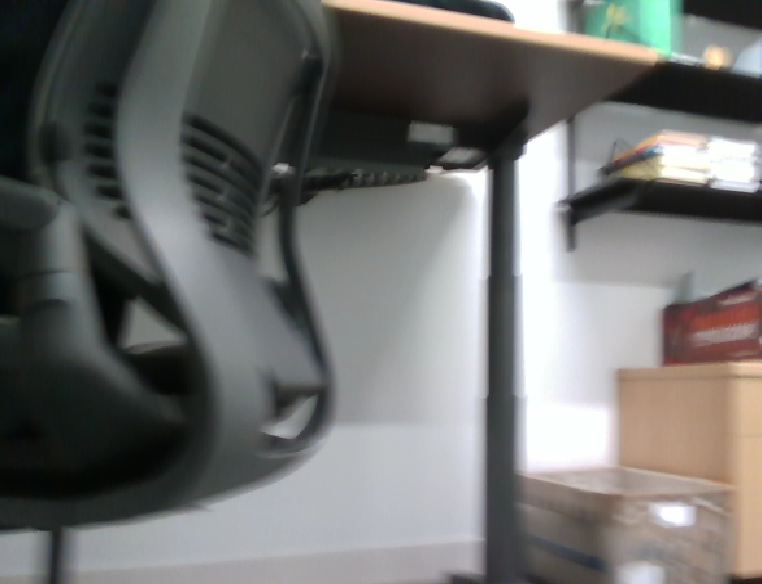}\\
\vspace*{0.2em}
\includegraphics[width=0.12\linewidth]{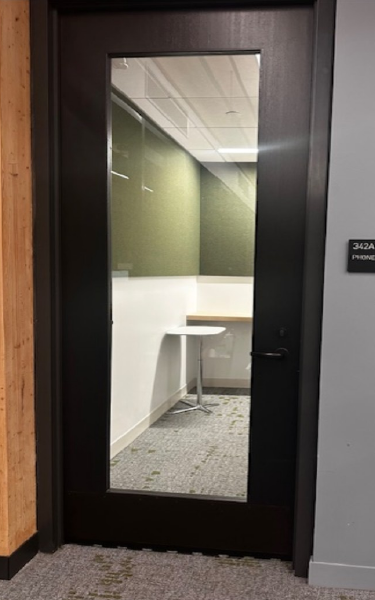}
\includegraphics[width=0.27\linewidth]{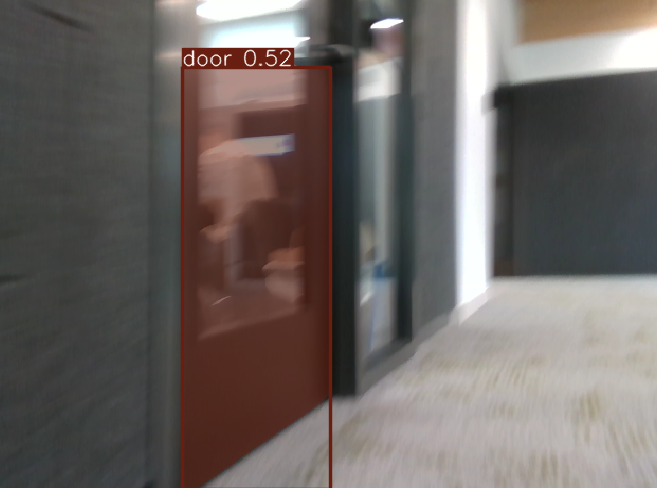}
\includegraphics[width=0.27\linewidth]{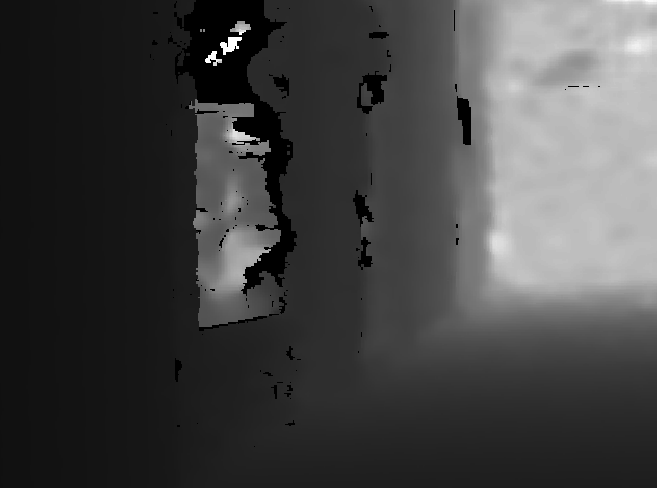}
\caption{(Top) The low vantage of the Go2 can create large occlusions and unusual viewpoints for chairs and tables. (Bottom) Examples of glass doors (left) in our data with a segmentation mask (middle) and depth estimate (right).}
\label{fig:door-table}
\end{figure}

\begin{remark}
From the vantage point of the Go2, tables are difficult to segment. Notice in~\cref{fig:segmentation} the missing detection of the table. Even when specifically prompted, detections of tables over the course the experiment are sparse. In the small office, the Go2 can only see the legs and only part of the underside of the table surface.
Doors are difficult for a different reason. In this office building, doors are paned with glass. See~\cref{fig:door-table}. In spite of this, the YOLOE model does segment them (measured by the number of predicted doors) but localization is difficult due to incorrect depth from the RealSense camera. We pad both the footprint of the door ground truth bounding box and the predicted bounding boxes by 0.5 m to help with quantitative analysis. We decided to include this semantic class even in the face of poor performance because we feel that for a robot, localizing doors is essential for operation in realistic indoor scenes.
\end{remark}

We report the computational inference times of the 4 methods in~\cref{tab:computationtime}. Clio is the fastest method defending its claim of real-time. Our online SAM3D-less method is quite fast despite having quite a few parts and it scales pretty well with multiple masks per frame. Including SAM3D can improve results but at the cost of larger inference times.

\begin{table}[!htpb]
    \centering
    \LARGE
    \resizebox{\linewidth}{!}{
    \renewcommand{\arraystretch}{1.5}
    \begin{tabular}{c|c|c|c|c|c|c}
    \toprule
    \textbf{Method} & \textbf{E2E per object (s)} & \textbf{E2E per frame (s)} & \textbf{Segmentation (s)} & \textbf{Retrieval (s)} & \textbf{Registration (s)} & \textbf{Reconciliation (s)}\\
    \midrule
        Clio & 0.025 $\pm$ 0.013 & 0.129 $\pm$ 0.045 & 0.004 $\pm$ 0.001 & N/A & N/A & 0.013 $\pm$ 0.008\\
        SAM3D & 23.417 $\pm$ 9.197 & 25.818 $\pm$ 9.504 & 0.497 $\pm$ 0.122 & N/A & N/A & N/A\\
        Ours & 1.603 $\pm$ 1.070 & 1.678 $\pm$ 1.048 & 0.128 $\pm$ 0.063 & 0.250 $\pm$ 0.069 & 0.252 $\pm$ 0.380 & 0.004 $\pm$ 0.003 \\ 
        SAM3D+Ours & 26.341 $\pm$ 10.088 & 28.817 $\pm$ 10.784 & 0.497 $\pm$ 0.122 & 2.234 $\pm$ 1.502 & 0.633 $\pm$ 0.111 & 0.008 $\pm$ 0.007 \\
    \bottomrule
    \end{tabular}
    }
    \caption{Computational time for each method. End to end (E2E) inference time per object is the time delay between recording an image and adding an object from it to the map (averaged across objects). Inference per frame denotes the time delay in recording the image and processing all masks within that image using our entire pipeline.}
    \label{tab:computationtime}
\end{table}

\paragraph{Qualitative Results}
We highlight the lounge experiment in~\cref{fig:qualitative_results} for qualitative analysis of our results. We plot the 2D predicted bounding boxes of each method with the associated label against the ground truth. Clio (top right) which attempts to bound similar CLIP features, tends to bound the free space in between semantically similar objects, i.e., a cluster of multiple chairs will have a large single bounding box. The SAM3D experiment (second row), occasionally creates large bounding boxes due badly generated objects, e.g. a mask captured the chair and some of the background for which SAM3D unsuccessfully attempts to generate an entire room. We observe that our method (third row) and SAM3D+Ours (bottom row) have smaller predicted bounding boxes which reflect the fact that predictions are grounded to known objects (former) and retrieval mitigates bad generations from SAM3D (latter). One can also observe the issues we have with glass in this experiment. The errant objects localized around (44, -31) are due to reflections off of the windows in the back of the room.

\subsection{Semantic Navigation Experiments}

\input{simulation-experiments}

\paragraph{Real World}
Finally, we show an example of semantic navigation in our hallway scene. The experiment goes as follows: (1) we teleoperate the Go2 around the area of interest building the metric-semantic map, (2) the resulting map is fed as text (USDA file) to Gemini along a task prompt. (3) After receiving the response from the LLM, the Go2 autonomously performs waypoint based path planning and control via Nav2~\cite{macenski2020marathon2} to move between waypoints.
{ \color{gray}
\begin{framed}
    \tiny
    \noindent \textbf{User}: We would like to know how many offices are connected to this hallway. The doors to the offices might not be provided in the data, where might they be? Provide some waypoints of where the robot should go for doors that might not have been seen. Note that waypoints should be given as (x,y).

        \noindent \textbf{Robot}:
1. Extracting Relevant Coordinates

The xformOp:translate values in the USD give us the (x, y, z) coordinates of the objects. We'll use the x and y components to define our waypoints. Here's how we can extract coordinates for some key objects to guide our search:

Doors: object\_301: (-0.5, -12.8), object\_50: (-6.3, -10.1), object\_214: (8.7, -16.9)
     
Seating Areas (to check nearby): object\_271: (13.4, -14.5), object\_160: (6.0, -12.5), object\_38: (-7.6, -6.9)
     
Tables (to check opposite walls): object\_158: (5.7, -12.7), object\_39: (-7.8, -5.9)\\

\noindent 2. Robot Search Waypoints (x, y)

Here's a set of waypoints for the robot to follow, incorporating the object locations and a systematic search pattern:\\
Start: (0, 0) - Assuming this is a reasonable origin point.\\
Check Door 1: (-0.5, -12.9) -  Slightly offset from object\_301.\\
Check Table Area: (-7.8, -6.0) -  Near object\_39.\\
Check Door 2: (-6.3, -10.1) - Slightly offset from object\_50.\\
Check Table Area 2: (5.7, -12.8) - Near object\_158.\\
Check Seating Area 1: (6.0, -12.5) - Near object\_160.\\
Check Seating Area 2: (13.4, -14.5) - Near object\_271.\\

Explanation of Strategy:
     We start at an origin.
     We visit the known door locations.
     We systematically traverse the hallway in a clockwise direction.
     We check areas near seating and tables, as these might be near entrances to other rooms.
This set of waypoints provides a comprehensive search pattern, increasing the likelihood of finding both known and unknown doors.
\end{framed}
}

Observe that Gemini can discern from our general task of searching for offices that it must extract locations of doors, seating areas (which are not explicitly defined but rather inferred by Gemini via clusters of chairs), and tables in the hallway scene via our USD in spite of it being rather noisy, see~\cref{fig:qualitative_results} (middle). It understands that to find offices, it should navigate along the hallway and check near clusters of furniture which could lead to finding offices.

%% file: simulation-experiments.tex

\paragraph{Simulation experiments using a known metric-semantic representation of the scene}
\label{sec:sim-exp}

Our goal in this experiment is to show that under the assumption that the robotics stack is in place, e.g. localization, planning, and waypoint control, that a written metric-semantic map, e.g. universal scene description (USD), when used as a contextual prompt to LLMs, is a sufficient representation for performing ambiguously-prescribed tasks in semantically rich environments across long time-horizons and large scenes that contain hundreds of unique assets. 

These simulation experiments focus on two scenes, a warehouse and a hospital floor. We use Isaac Sim and MobilityGen~\cite{MobilityGen} to work with sensor streams (RGB-D and pose) and control commands of simulated Unitree H1 humanoid and Boston Dynamics Spot quadruped robots. Isaac Sim natively uses Universal Scene Description (USD) \cite{OpenUSD} to represent scenes. A USD is an XML-based hierarchical composition of assets developed by Pixar as a language to store assets, construct scenes, and even compose movies. In its ASCII form (.usda), it is a human-readable text file format which can be read by LLMs and humans alike or converted into knowledge graphs~\cite{nguyen2023translatinguniversalscenedescriptions}. Given the ground-truth USD of the scene, which inherently is a metric-semantic map which relates assets and its labels to their metric pose, the goal is to use an LLM (Gemini via an API call) to generate a list of waypoints (position and orientation) in the global frame, and use classical planning and control to navigate the robot.

\begin{wrapfigure}{r}{0.5\linewidth}
{
\color{gray}
\begin{framed}
    \tiny
    \noindent \textbf{User}: Plan a path of waypoints that navigates you around the hospital floor. Waypoints should fall within the main hospital walls but not within ANY of the objects in the USD. Be sure to choose waypoints that view some interesting objects and areas.

        \noindent \textbf{Robot}:
        Description: Planned path around the hospital floor, including interesting objects and equipment along the route. \\
        1. Start near the entrance, view the reception desks.\\
        2. Move towards the water dispenser and pharmacy area.\\
        3. Navigate near the chairs, side tables and scale.\\
        4. Head towards the area with multiple shelves and supply carts.\\
        5. Move to a larger area with water dispenser, side table and trashcan.\\
        6. Navigate towards the gurney and medical bags.\\
        7. Final waypoint.
\end{framed}
}
\end{wrapfigure}
\begin{figure} [!htpb]
    \centering
    \includegraphics[width=0.49\linewidth]{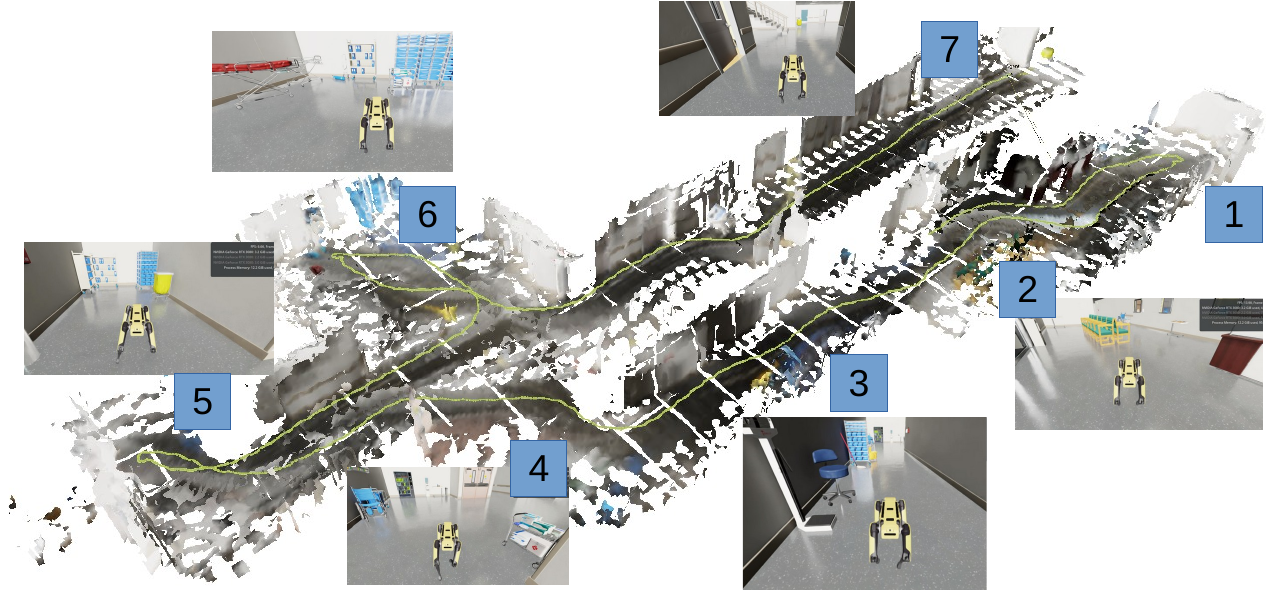}
    \hspace*{0.2em}
    \includegraphics[width=0.41\linewidth]{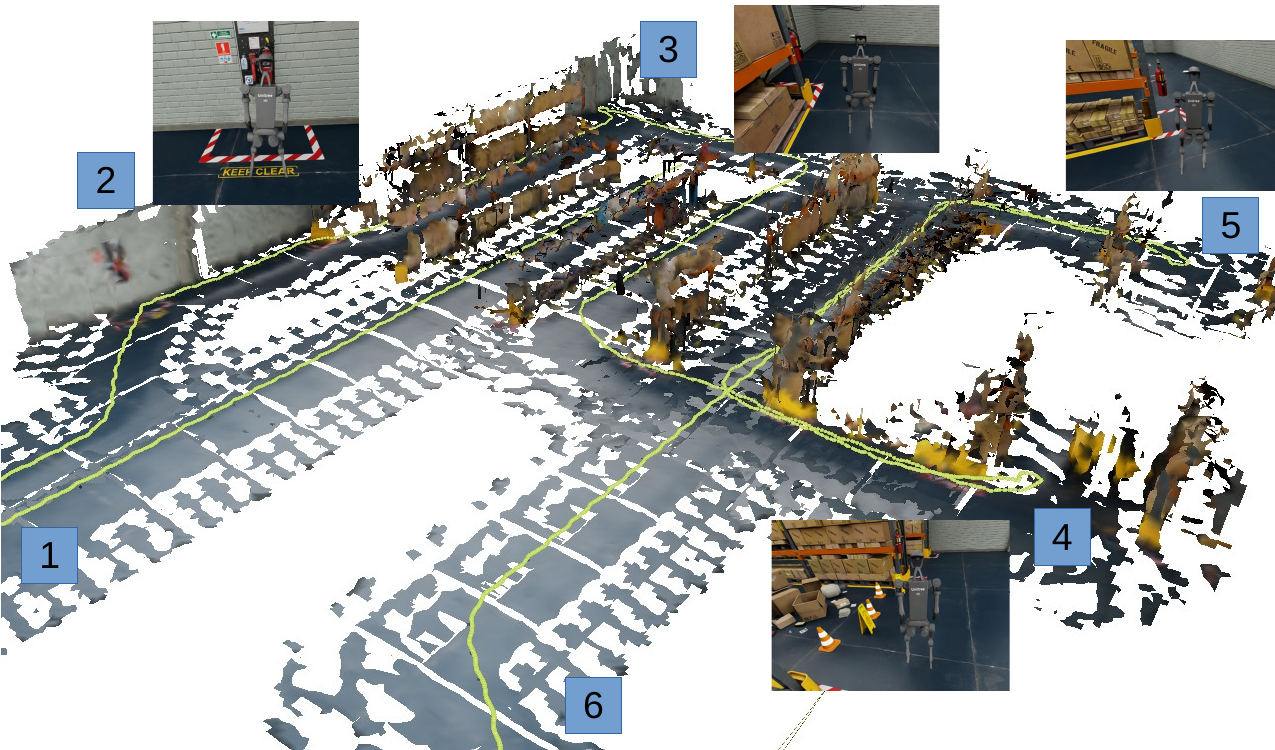}
    \caption{(Left) Snapshots of the Spot robot at different waypoints in the hospital scene. The 3D mesh was built from simulated RGB, depth, and pose data using Voxblox~\cite{oleynikova2017voxblox} to demonstrate coverage. (Right) Inset images show safety equipment such as fire extinguishers and wet floor signs that were discovered by a Unitree H1 robot exploring the warehouse scene.}
    \label{fig:sim-experiments}
\end{figure}

\cref{fig:sim-experiments} (left) shows an example where Spot navigates a hospital floor that contains 55 unique objects (1756 instances) ranging from hospital beds, soda machines, and assorted furniture (46 categories) to walls and floor descriptions (9 types). The prompt and the textual description of the waypoints are shown in the adjoining box. We have not shown the 2D Euclidean coordinates returned by Gemini that are used for waypoint-based planning. Spot travels a distance of 128 \unit{m} during this experiment with minimal backtracking and traverses a majority of the scene. Approximately 81\% of the scene is observed (2371 out of 2922 \unit{m^2}).

With essentially the same prompt (not shown for lack of space), the simulated Unitree H1 robot can search for things in a warehouse that could be useful in an emergency, see~\cref{fig:sim-experiments} (right). The warehouse scene is approximately 1687 \unit{m^2} with 45 unique objects such as a forklift, boxes and crates. The robot obtains data from all 4 corners of the scene but does not enter some of the aisles. In this experiment, the robot traveled a distance of 174.1 \unit{m} with substantial amounts of backtracking. Reordering waypoints given by Gemini would reduce this path length.


%% file: conclusion.tex

\section{Discussion}
\label{sec:discussion}

In this work, we suggest that if the most natural way for a human to provide instruction is through language, our robots must be able to interpret these natural language specifications within the context of the semantic and metric information in the environment.
A natural procedure to instantiate this paradigm is to build a scene-level representation of the scene that is geometrically and semantically accurate.
Such representations, if stored in JSON or USDA, can also be easily read by LLMs to result in a powerful framework for precise language-based task specifications.
This paper demonstrated a general pipeline to build metric-semantic representations of indoor scenes from robot sensor data to
perform complex language-guided tasks in simulation, as well as on a quadruped Unitree Go2.
In our opinion, the key challenges for such indoor navigation tasks that one must focus on in future work pertain to (i) improving the robustness and generalization of deep learning-based methods such as YOLO, SAM or SAM3D to real world data which has motion blur from being mounted on a walking platform, illumination and viewpoint changes, as well as nuisances such as glass or shiny surfaces, and (ii) addressing the inevitable computational and communication latency that comes from combining large deep networks into real robotics pipelines.